\documentclass{article}

\usepackage{graphicx}
\usepackage{xcolor}
\usepackage{hyperref}
\usepackage{bm}
\usepackage{url}
\usepackage{booktabs}
\usepackage{tablefootnote}
\usepackage{amsmath,array,amsfonts}
\usepackage{pifont}
\usepackage{xspace}
\usepackage{listings}
\usepackage{tcolorbox,cuted}
\usepackage{wrapfig}
\usepackage{subfigure}
\usepackage[bottom]{footmisc}

\usepackage{hyperref}

\usepackage[accepted]{icml2020}

\def\e{{\mathbf e}}

\def\x{{\mathbf x}}

\def\A{{\mathbf A}}

\def\E{{\mathbf E}}

\def\X{{\mathbf X}}

\def\cG{\mathcal{G}}  
\def\cX{\mathcal{X}}  
\def\cE{\mathcal{E}}  

\definecolor{mGreen}{rgb}{0.282, 0.635, 0.247}
\definecolor{mPurp}{rgb}{0.4940, 0.1840, 0.5560}
\definecolor{mOrange}{rgb}{0.9, 0.56, 0.28}

\newcommand{\ie}{\emph{i.e.}\xspace}

\renewcommand{\paragraph}[1]{\textbf{#1\;}}

\definecolor{ForestGreen}{rgb}{0.0, 0.57, 0.13}

\newcommand{\docs}{\url{https://graphneural.network}}
\newcommand{\github}{\url{https://github.com/danielegrattarola/spektral}}

\lstdefinestyle{custom}{
  belowcaptionskip=1\baselineskip,
  breaklines=true,
  frame=none,
  xleftmargin=\parindent,
  language=python,
  showstringspaces=false,
  basicstyle=\ttfamily,
  keywordstyle=\bfseries\color{green!40!black},
  commentstyle=\itshape\color{purple!40!black},
  identifierstyle=\color{blue},
  stringstyle=\color{orange},
}
\renewcommand{\paragraph}[1]{\textbf{#1\;}}

\begin{document}

\twocolumn[
\icmltitle{Graph Neural Networks in TensorFlow and Keras with Spektral}
\begin{icmlauthorlist}
\icmlauthor{Daniele Grattarola}{usi}
\icmlauthor{Cesare Alippi}{usi,poli}
\end{icmlauthorlist}
\icmlaffiliation{usi}{Universit\`{a} della Svizzera italiana, Lugano, Switzerland}
\icmlaffiliation{poli}{Politecnico di Milano, Milan, Italy}
\icmlcorrespondingauthor{Daniele Grattarola}{grattd@usi.ch}
\icmlkeywords{Graph Neural Networks, Graph Deep Learning, TensorFlow, Keras}
\vskip 0.3in
]
\printAffiliationsAndNotice{}

\begin{abstract}
    In this paper we present Spektral, an open-source Python library for building graph neural networks with TensorFlow and the Keras application programming interface. 
    Spektral implements a large set of methods for deep learning on graphs, including message-passing and pooling operators, as well as utilities for processing graphs and loading popular benchmark datasets. 
    The purpose of this library is to provide the essential building blocks for creating graph neural networks, focusing on the guiding principles of user-friendliness and quick prototyping on which Keras is based. Spektral is, therefore, suitable for absolute beginners and expert deep learning practitioners alike. 
    In this work, we present an overview of Spektral's features and report the performance of the methods implemented by the library in scenarios of node classification, graph classification, and graph regression. 
\end{abstract}

\section{Introduction}

Graph Neural Networks (GNNs) are a class of deep learning methods designed to perform inference on data described by graphs \cite{battaglia2018relational}. 
Due to the different possibilities offered by graph machine learning and the large number of applications where graphs are naturally found, GNNs have been successfully applied to a diverse spectrum of fields to solve a variety of tasks. In physics, GNNs have been used to learn physical models of complex systems of interacting particles \cite{battaglia2016interaction,kipf2018neural,sanchez2018graph,farrell2018novel}. In recommender systems, the interactions between users and items can be represented as a bipartite graph and the goal is to predict new potential edges (\ie, which items could a user be interested in), which can be achieved with GNNs \cite{berg2017graph,ying2018graph}. GNNs have also been largely applied to the biological sciences, with applications ranging from the recommendation of medications \cite{shang2019gamenet}, to the prediction of protein-protein and protein-ligand interactions \cite{gainza2020deciphering}, and in chemistry, for the prediction of quantum molecular properties as well as the generation of novel compounds and drugs \cite{do2019graph,you2018graphrnn}.
Finally, GNNs have been successfully applied in fields like natural language processing \cite{fernandes2018structured,de2018question} and even more complex tasks like abstract reasoning \cite{santoro2017simple,allamanis2017learning,schlichtkrull2018modeling} and decision making with reinforcement learning \cite{zambaldi2018relational,hamrick2018relational}.

\begin{figure*}[]
    \centering
    \subfigure[]{\includegraphics[width=0.775\textwidth]{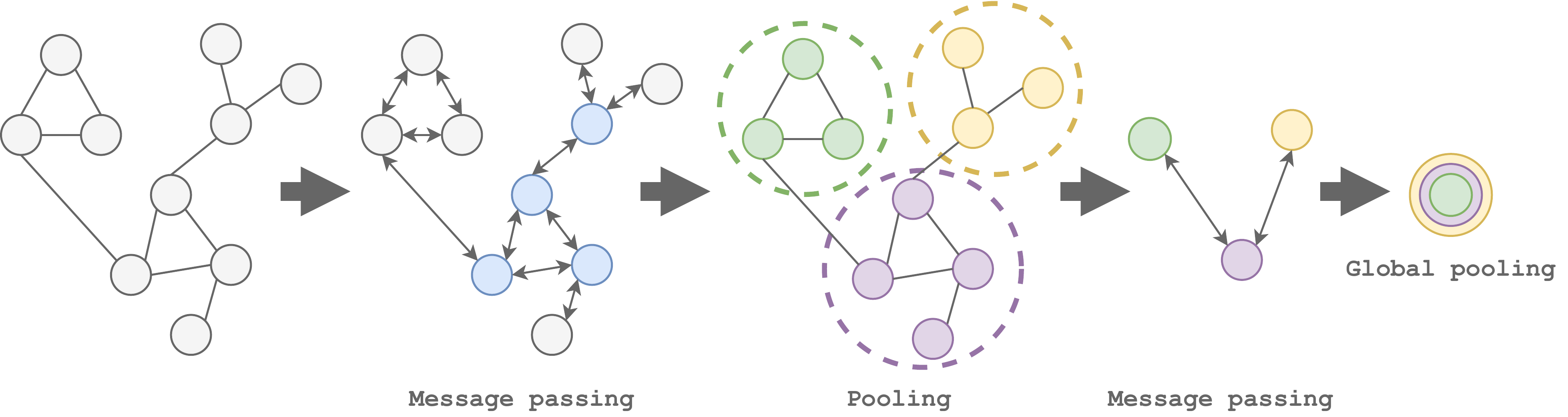}\label{fig:gnn_1}}\hfill
    \subfigure[]{\includegraphics[width=0.175\textwidth]{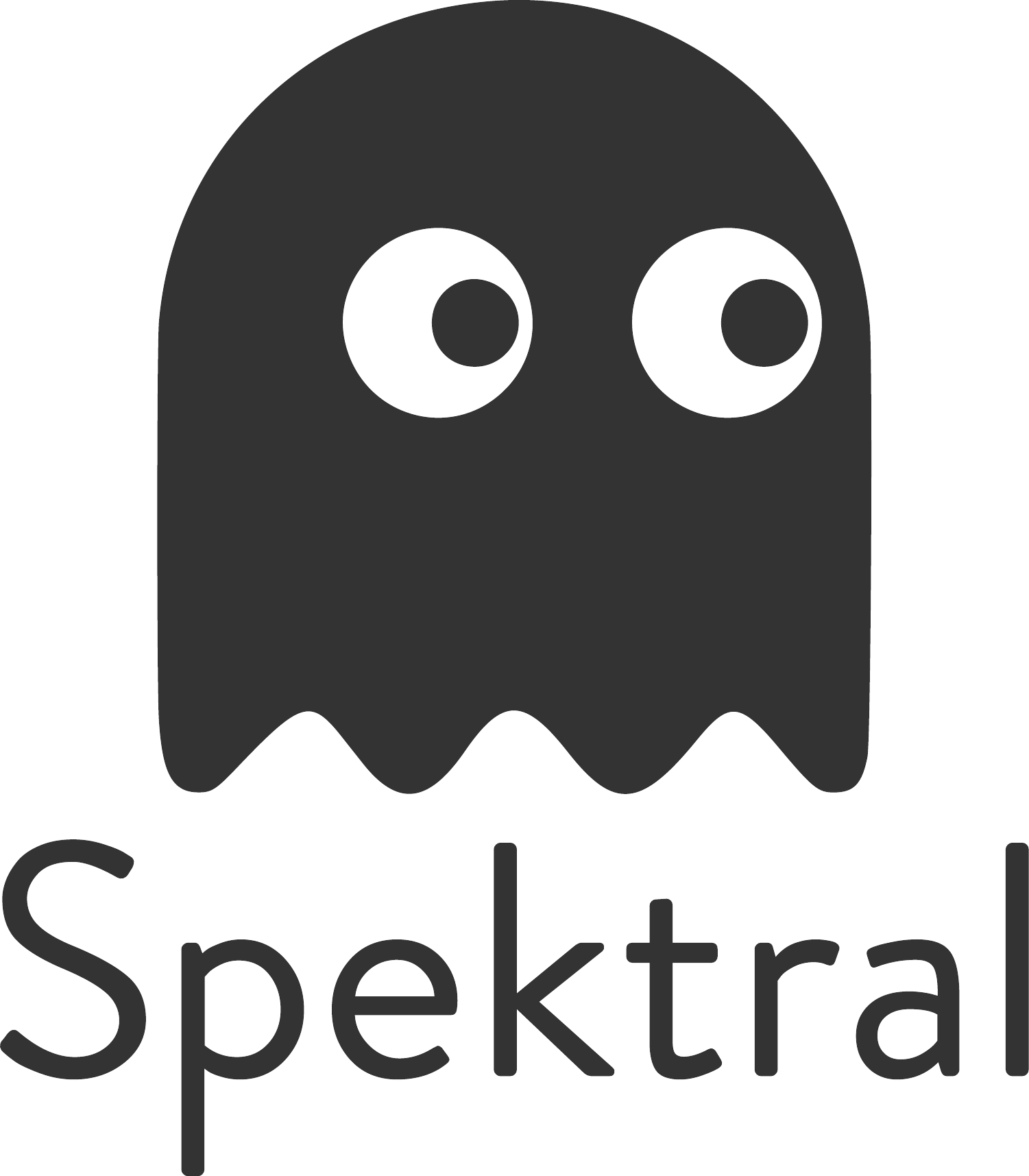}\label{fig:gnn_2}}
    \caption{\protect\ref{fig:gnn_1} Schematic view of a graph neural network with message-passing, pooling, and global pooling layers. The role of message-passing layers is to compute a representation of each node in the graph, leveraging \textit{local} information (\textit{messages}) from its neighbours. The role of pooling layers is to reduce the size of the graph by aggregating or discarding redundant information, so that the GNN can learn a hierarchical representation of the input data. Finally, global pooling layers reduce the graph to a single vector, usually to feed it as input to a multi-layer perceptron for classification or regression.
    \protect\ref{fig:gnn_2} The stylised ghost logo of Spektral.}
    \label{fig:gnn}
\end{figure*}

At the core of GNNs there are two main types of operations, which can be interpreted as a generalisation of the convolution and pooling operators in convolutional neural networks: \textit{message passing} and \textit{graph pooling} (Fig.\ \ref{fig:gnn}). The former is used to learn a non-linear transformation of the input graphs and the latter to reduce their size. 
When combined, these two operations enable graph representation learning as a general tool to predict node-level, edge-level, and global properties of graphs.
Several works in recent literature have introduced models for either message passing \cite{gilmer2017neural,scarselli2009graph,defferrard2016convolutional,kipf2016semi,simonovsky2017dynamic,velickovic2017graph,hamilton2017inductive,bianchi2019graph,xu2018powerful,klicpera2019predict} or graph pooling \cite{ying2018hierarchical,graphunet,bianchi2019mincut,cangea2018towards}. 

Thanks to the increasing popularity of GNNs, many software libraries implementing the building blocks of graph representation learning have been developed in recent years, paving the way for the adoption of GNNs in other fields of science. One of the major challenges faced by researchers and software developers who wish to contribute to the larger scientific community is to make software both accessible and intuitive, so that even non-technical audiences can benefit from the advances carried by intelligent systems. 
In this spirit, Keras is an application programming interface (API) for creating neural networks, developed according to the guiding principle that ``being able to go from idea to result with the least possible delay is key to doing good research"~\cite{chollet2015keras}. Keras is designed to reduce the cognitive load of end users, shifting the focus away from the boilerplate implementation details and allowing instead to focus on the creation of models. As such, Keras is extremely beginner-friendly and, for many, an entry point to machine learning itself. 
At the same time, Keras integrates smoothly with its TensorFlow \cite{abadi2016tensorflow} backend and enables users to build any model that they could have implemented in pure TensorFlow. This flexibility makes Keras an excellent tool even for expert deep learning practitioners and has recently led to TensorFlow's adoption of Keras as the official interface to the framework. 

In this paper we present Spektral, a Python library for building graph neural networks using TensorFlow and the Keras API. Spektral implements some of the most important papers from the GNN literature as Keras layers, and it integrates seamlessly within Keras models and with the most important features of Keras like the training loop, callbacks, distributed training, and automatic support for GPUs and TPUs. As such, Spektral inherits the philosophy of ease of use and flexibility that characterises Keras. The components of Spektral act as standard TensorFlow operations and can be easily used even in more advanced settings, integrating tightly with all the features of TensorFlow and allowing for an easy deployment to production systems. For these reasons, Spektral is the ideal library to implement GNNs in the TensorFlow ecosystem, both for total beginners and experts alike.

All features of Spektral are documented in detail\footnote{\docs} and a collection of examples is provided with the source code. 
The project is released on GitHub\footnote{\github} under MIT license. 

\section{Library Overview}

\subsection{Representing graphs} 
\label{sec:graphs}

Let $\cG = \{\cX, \cE\}$ be a graph where $\cX = \{ \x_i \in \mathbb{R}^F | i = 1, \dots, N \}$ is the set of nodes with $F$-dimensional real attributes, and $\cE = \{\e_{ij} \in \mathbb{R}^S | \x_i, \x_j \in \cX \}$ the set of edges with $S$ dimensional real attributes.
In Spektral, we represent $\cG$ by its binary adjacency matrix $\A \in \{0, 1\}^{N \times N}$, node features $\X \in \mathbb{R}^{N \times F}$, and edge features $\E \in \mathbb{R}^{N \times N \times S}$.
Any format accepted by Keras to represent the above matrices is also supported by Spektral, which means that it also natively supports the NumPy stack of scientific computing libraries for Python. Most of the layers and utilities implemented in Spektral also support the sparse matrices of the SciPy library, making them computationally efficient both in time and memory. Additionally, Spektral makes very few assumptions on how a user may want to represent graphs, and transparently deals with batches of graphs represented as higher-order tensors or disjoint unions (see Appendix \ref{app:data-modes} for more details).

\subsection{Message passing} 

Message-passing networks are a general paradigm introduced by \citet{gilmer2017neural} that unifies most GNN methods found in the literature as a combination of \textit{message}, \textit{aggregation}, and \textit{update} functions. Message-passing layers are equivalent in role to the convoutional operators in convolutional neural networks, and are the essential component of graph representation learning. 
Message-passing layers in Spektral are available in the \lstinline{layers.convolutional} module.\footnote{The name \textit{convolutional} derives from the homonymous module in Keras, as well as message-passing layers being originally derived as a generalisation of convolutional operators.} Currently, Spektral implements fifteen different message-passing layers including Graph Convolutional Networks (GCN) \cite{kipf2016semi}, ChebNets \cite{defferrard2016convolutional}, GraphSAGE \cite{hamilton2017inductive}, ARMA convolutions \cite{bianchi2019graph}, Edge-Conditioned Convolutions (ECC) \cite{simonovsky2017dynamic}, Graph Attention Networks (GAT) \cite{velickovic2017graph}, APPNP \cite{klicpera2019predict}, and Graph Isomorphism Networks (GIN) \cite{xu2018powerful}, as well as the methods proposed by \citet{li2017diffusion,li2015gated,thekumparampil2018attention,du2017topology,xie2018crystal,wang2018dynamic} and a general interface that can be extended to implement message-passing layers. The available methods are sufficient to deal with all kinds of graphs, including those with attributed edges.

\subsection{Graph pooling} 

Graph pooling refers to any operation to reduce the number of nodes in a graph and has a similar role to pooling in traditional convolutional networks for learning hierarchical representations. Because pooling computes a coarser version of the graph at each step, ultimately resulting in a single vector representation, it is usually applied to problems of graph-level inference.
Graph pooling layers are available in \lstinline{layers.pooling} and include: DiffPool \cite{ying2018hierarchical}, MinCut pooling \cite{bianchi2019mincut}, Top-$K$ pooling \cite{cangea2018towards,graphunet}, and Self-Attention Graph Pooling (SAGPool) \cite{lee2019self}.
Spektral also implements \textit{global} graph pooling methods, which can be seen as a limit case of graph pooling where a graph is reduced to a single node, \emph{i.e.}, its node features are reduced to a single vector. Spektral implements six different global pooling strategies: sum, average, max, gated attention (GAP) \cite{li2015gated}, SortPool \cite{zhang2018end}, and attention-weighted sum (AWSP).\footnote{While never published in the literature, attention-weighted sum is a straightforward concept that consists of computing a weighted sum of the node features, where the weights are computed through a simple attentional mechanism}

\subsection{Datasets}

Spektral comes with a large variety of popular graph datasets accessible from the \lstinline{datasets} module. The datasets available from Spektral provide benchmarks for transductive and inductive node classification, graph signal classification, graph classification, and graph regression. In particular, the following datasets can be loaded with Spektral: the citation networks, Cora, CiteSeer, and Pubmed \cite{sen2008collective}; the protein-protein interaction dataset (PPI)~\cite{stark2006biogrid,zitnik2017predicting,hamilton2017inductive} and the Reddit communities network dataset \cite{hamilton2017inductive} from the GraphSAGE paper \cite{hamilton2017inductive}; the QM9 chemical dataset of small molecules \cite{ramakrishnan2014quantum}; the MNIST 8-NN graph for graph signal classification as proposed by \citet{defferrard2016convolutional}; the Benchmark Data Sets for Graph Kernels \cite{kersting2016benchmark}.
Each dataset is automatically downloaded and stored locally when necessary.

\subsection{Other tools} 

Some of the secondary functionalities implemented by Spektral include: the \lstinline{utils} module, which exposes some useful utilities for graph deep learning (\emph{e.g.}, methods for computing the characteristic graph matrices or manipulating the data); the \lstinline{chem} module, which offers tools for loading and processing molecular graphs; the \lstinline{layers.ops} module, which offers a set of common operations that can be used by advanced users to create new GNN layers, like wrappers for common matrix operations that automatically handle sparse inputs and batches of graphs.

\begin{table}[]
    \centering
    \caption{Comparison of different GNN libraries. The \emph{Framework} column indicates the backend framework supported by the library, while the \textit{MP} and \textit{Pooling} columns indicate the number of different message-passing and pooling layers implemented by the library, respectively. }
    \begin{tabular}{llcc}
        \toprule
        \textbf{Library} &  \textbf{Framework} & \textbf{MP} & \textbf{Pooling} \\
        \midrule
        \textbf{Spektral} & TensorFlow & 15 & 10 \\
        \textbf{PyG} & PyTorch & 28 & 14\\
        \textbf{DGL} & PyTorch, MXNet & 15 & 7 \\
        \textbf{StellarGraph} & TensorFlow & 6 & N/A \\
        \bottomrule
    \end{tabular}
    \label{tab:frameworks}
\end{table}

\section{Comparison to other libraries}

Given the growing popularity of the field, several libraries for GNNs have appeared in recent years. Among the most notable, we cite PyTorch Geometric\footnote{\url{https://pytorch-geometric.readthedocs.io/}} (PyG) \cite{fey2019fast} and the Deep Graph Library\footnote{\url{https://docs.dgl.ai/}} (DGL) \cite{wang2019deep}, both of which are based on the PyTorch deep learning library.
Instead, Spektral is developed for the TensorFlow ecosystem, which to this date is estimated to support the majority of deep learning applications both in research and industry \cite{2019why}. 
The features offered by Spektral, summarized in Table \ref{tab:frameworks}, are largely similar to those offered both by PyG (which however implements a much larger variety of message-passing methods and other algorithms from GNN literature) and by DGL. 
The computational performance of Spektral's layers is also comparable to that of PyG, with small differences due to implementation details and differences between the two respective backend frameworks.
We also mention the StellarGraph library for GNNs which, like Spektral, is based on Keras. This library implements six message-passing layers, four of which are available in Spektral (GCN, GraphSAGE, GAT and APPNP), but does not offer pooling layers and relies on a custom format for graph data, which limits flexibility. 

\section{Applications}
\label{sec:experiments}

In this section, we report some experimental results on several well-known benchmark tasks, in order to provide a high-level overview of how the different methods implemented by Spektral perform in a standard research use case scenario. We report the results for three main settings: a node classification task and two tasks of graph-level property prediction, one of classification and one of regression.
All experimental details are reported in Appendix \ref{app:A}.

\paragraph{Node classification} 
In our first experiment, we consider a task of semi-supervised node classification on the Cora, CiteSeer, and Pubmed citation networks. 
We evaluate GCN \cite{kipf2016semi}, ChebNets \cite{defferrard2016convolutional}, ARMA \cite{bianchi2019graph}, GAT \cite{velickovic2017graph} and APPNP \cite{klicpera2019predict}.
We reproduce the same experimental settings described in the original papers, but we use the random data splits suggested by \citet{shchur2018pitfalls} for a fairer evaluation.

\paragraph{Graph classification} 
\label{sec:gc}
To evaluate the pooling layers, DiffPool, MinCut, Top-$K$, and SAGPool, we consider a task of graph-level classification, where each graph represents an individual sample to be classified. We use four datasets from the Benchmark Data Sets for Graph Kernels: Proteins, IMDB-Binary, Mutag and NCI1.
Here, we adopt a fixed GNN architecture (described in Appendix \ref{app:A}) where we only change the pooling method. To assess whether each pooling layer is actually beneficial for learning a representation of the data, we also evaluate the same GNN without pooling (\textit{Flat}).

\paragraph{Graph regression} 
To evaluate the global pooling methods, we use the QM9 molecular database and train a GNN on four different regression targets: dipole moment (Mu), isotropic polarizability (Alpha), energy of HOMO (Homo), and internal energy at OK (U0). Because the molecules in QM9 have attributed edges, we adopt a GNN based on ECC, which is designed to integrate edge attributes in the message-passing operation.

\subsection{Results}

The results for each experiment are reported in Tables \ref{tab:nc}, \ref{tab:gc}, and \ref{tab:gr}. In the first experiment, results are compatible with what reported in the literature, although some differences in performance are present due to using a random data split rather than the pre-defined one used in the original experiments. The APPNP operator consistently achieves good results on the citation networks, outperforming the other methods on Cora and Pubmed, and coming close to ARMA on CiteSeer. 
For graph classification, the results are sometimes different than what reported in the literature, due to the standardized architecure that we used in this experiment. MinCut generally achieves the best performance followed by DiffPool. We also note that the Flat baseline often achieves better results than the equivalent GNNs equipped with pooling.
For graph regression, results show that the choice of global pooling method can have a significant impact on performance. In particular, the GAP operator performs best on Alpha and U0, while the best results on Mu and Homo are obtained with max pooling and AWSP, respectively. However, we note how these latter two operators have largely unstable performances depending on the datasets, as both fail on Alpha and U0.

\begin{table}
    \centering
        \setlength{\tabcolsep}{0.3em}
        \centering
        \caption{Classification accuracy on the node classification tasks.}
        \resizebox{\columnwidth}{!}{%
        \begin{tabular}{lcccccccc}
            \toprule
            \textbf{Dataset} & \textbf{ChebNets} & \textbf{GCN} & \textbf{GAT} & \textbf{ARMA} & \textbf{APPNP} \\
            \midrule
            \textbf{Cora}     & 77.4 \tiny{$\pm 1.5$} & 79.4 \tiny{$\pm 1.3$} & 82.0 \tiny{$\pm 1.2$} & 80.5 \tiny{$\pm $} & \textbf{82.8 \tiny{$\pm 0.9$}} \\
            \textbf{Citeseer} & 68.2 \tiny{$\pm 1.6$} & 68.8 \tiny{$\pm 1.4$} & 70.0 \tiny{$\pm 1.0$} & \textbf{70.6 \tiny{$\pm 0.9$}} & 70.0 \tiny{$\pm 1.0$} \\
            \textbf{Pubmed}   & 74.0 \tiny{$\pm 2.7$} & 76.6 \tiny{$\pm 2.5$} & 73.8 \tiny{$\pm 3.3$} & 77.2 \tiny{$\pm 1.6$} & \textbf{78.2 \tiny{$\pm 2.1$}} \\
            \bottomrule
        \end{tabular}
        }
        \label{tab:nc}
\end{table}

\begin{table}
        \setlength{\tabcolsep}{0.3em}
        \centering
        \caption{Classification accuracy on the graph classification tasks.}
        \resizebox{\columnwidth}{!}{%
        \begin{tabular}{lccccc}
            \toprule
            \textbf{Dataset}   & \textbf{Flat}          & \textbf{MinCut}       & \textbf{DiffPool}     & \textbf{Top-$K$}          & \textbf{SAGPool} \\
            \midrule
            \textbf{Proteins}  & 74.3 \tiny{$\pm 4.5$}  & \textbf{75.5 \tiny{$\pm 2.0$}} & 74.1 \tiny{$\pm 3.9$} & 70.5 \tiny{$\pm 3.4$}  & 71.3 \tiny{$\pm 5.0$} \\
            \textbf{IMDB-B}    & 72.8 \tiny{$\pm 7.2$}  & \textbf{73.6 \tiny{$\pm 5.4$}} & 70.6 \tiny{$\pm 6.6$} & 67.7 \tiny{$\pm 8.2$}  & 69.3 \tiny{$\pm 5.7$} \\
            \textbf{Mutag}     & 72.5 \tiny{$\pm 14.0$} & 81.4 \tiny{$\pm 10.7$} & \textbf{83.5 \tiny{$\pm 9.7$}} & 79.2 \tiny{$\pm 8.0$} & 78.5 \tiny{$\pm 8.3$} \\
            \textbf{NCI1}      & \textbf{77.3 \tiny{$\pm 2.6$}}  & 74.4 \tiny{$\pm 1.9$} & 71.1 \tiny{$\pm 3.0$} & 72.0 \tiny{$\pm 3.0$}  & 69.4 \tiny{$\pm 8.4$} \\
            \bottomrule
        \end{tabular}}
        \label{tab:gc}
\end{table}  

\begin{table}
        \setlength{\tabcolsep}{0.3em}
        \centering
        \caption{Mean-squared error on the graph regression tasks. Results for Homo are in scale of $10^{-5}$.}
        \resizebox{\columnwidth}{!}{%
        \begin{tabular}{lccccc}
            \toprule
            \textbf{Dataset} & \textbf{Average} & \textbf{Sum} & \textbf{Max} & \textbf{GAP} & \textbf{AWSP} \\
            \midrule
            \textbf{Mu} & 1.12 \tiny{$\pm 0.03$} & 1.02 \tiny{$\pm 0.02$} & \textbf{0.90 \tiny{$\pm 0.04$}} & 1.04 \tiny{$\pm 0.05$} & 0.99 \tiny{$\pm 0.03$} \\
            \textbf{Alpha} & 3.15 \tiny{$\pm 0.65$} & 2.38 \tiny{$\pm 0.64$} & 6.20 \tiny{$\pm 0.33$} & \textbf{1.89 \tiny{$\pm 0.59$}} & 31.1 \tiny{$\pm 0.37$} \\
            \textbf{Homo} & 9.24 \tiny{$\pm 0.41$} & 9.22 \tiny{$\pm 0.51$} & 8.90 \tiny{$\pm 0.36$} & 9.04 \tiny{$\pm 0.29$} & \textbf{8.05 \tiny{$\pm 0.29$}} \\
            \textbf{U0} & 0.42 \tiny{$\pm 0.14$} & 0.50 \tiny{$\pm 0.13$} & 110.7 \tiny{$\pm 4.5$} & \textbf{0.22 \tiny{$\pm 0.13$}} & 624.0 \tiny{$\pm 19.0$} \\
            \bottomrule
        \end{tabular}}
        \label{tab:gr}
\end{table}

\section{Conclusion}

We presented Spektral, a library for building graph neural networks using the Keras API. Spektral implements several state-of-the-art methods for GNNs, including message-passing and pooling layers, a wide set of utilities, and comes with many popular graph datasets. 
The library is designed for providing a streamlined user experience and is currently the most mature library for GNNs in the TensorFlow ecosystem. In the future, we will keep Spektral up to date with the ever-growing field of GNN research, and we will focus on improving the performance of its core components. 

\bibliography{main}
\bibliographystyle{icml2020}

\clearpage
\appendix

\section{Data modes} 
\label{app:data-modes}
Spektral supports four different ways of representing graphs (or batches thereof), which we refer to as \textit{data modes}.

In \textit{single mode}, the data describes a single graph with its adjacency matrix and attributes, and inference usually happens at the level of individual nodes. \textit{Disjoint mode} is a special case of single mode, where the graph is obtained as the disjoint union of a set of smaller graphs. In this case the node attributes of the graphs are stacked in a single matrix and their adjacency matrices are combined in a block-diagonal matrix. This is a practical way of representing batches of variable-order graphs, although it requires an additional data structure to keep track of the different graphs is a batch. Alternatively, in \textit{batch mode}, a set of graphs is represented by stacking their adjacency and attributes matrices in higher-order tensors of shape $B \times N \times \dots$. This mode is akin to traditional batch processing in machine learning and can be more naturally adopted in deep learning architectures. However, it requires the graphs in a batch to have the same number of nodes. 
Finally, \textit{mixed mode} is the one most often found in traditional machine learning literature and consists of a graph with fixed support but variable attributes. Common examples of this mode are found in computer vision, where images have a fixed 2-dimensional grid support and only differ in the pixel values (\ie, the node attributes), and in traditional graph signal processing applications.

In Spektral, all layers implementing message-passing operations are compatible with single and disjoint mode, and more than half of the layers also support mixed and batch mode. A similar consideration holds for pooling methods where however, due to structural limits of the methods themselves, some data modes cannot be supported directly. 

\section{Technical notes}

Spektral is distributed through the Python Package Index (package name: \texttt{spektral}), supports all UNIX-like platforms,\footnote{It is also largely compatible with Windows.} and has no proprietary dependencies. The library is compatible with Python version 3.5 and above. 
Starting from version 0.2, Spektral is developed for TensorFlow 2 and its integrated implementation of Keras. Version 0.1 of Spektral, which is based on TensorFlow 1 and the stand-alone version of Keras, will be maintained until TensorFlow 1 is officially discontinued, although new features will only be added to the newer versions of Spektral. 

\section{Experimental details}
\label{app:A}
This section summarises the architectures and hyperparameters used in the experiments of Section \ref{sec:experiments}.
All experiments were run on a single NVIDIA Titan V GPU with 12GB of video memory.

\subsection{Node classification}

\noindent Hyperparameters:
\begin{itemize}
    \item Learning rate: see original papers;
    \item Weight decay: see original papers;
    \item Epochs: see original papers;
    \item Patience: see original papers;
    \item Repetitions per method and per dataset: 100;
    \item Data: we used Cora, Citeseer and Pubmed. As suggested in \cite{shchur2018pitfalls}, we use random splits with 20 labels per class for training, 30 labels per class for early stopping, all the remaining labels for testing. 
\end{itemize}

\subsection{Graph classification}

\noindent We configure a GNN with the following structure: \textsc{GCS - Pooling - GCS - Pooling - GCS - GlobalSumPooling - Dense},
where \textsc{GCS} indicates a \textit{Graph Convolutional Skip} layer as described in \cite{bianchi2019graph}, \textsc{Pooling} indicates the graph pooling layer being tested, \textsc{GlobalSumPooling} represents a global sum pooling layer, and \textsc{Dense} represents the fully-connected output layer.
GCS layers have 32 units each, ReLU activation, and L2 regularisation applied to both weight matrices. The same L2 regularization is applied to pooling layers when possible. Top-$K$ and SAGPool layers are configured to output half of the nodes for each input graph. DiffPool and MinCut are configured to output $K = \frac{\bar N}{2}$ nodes at the first layer, and $K = \frac{\bar N}{4}$ nodes at the second layer, where $\bar N$ is the average order of the graphs in the dataset. 
When using DiffPool, we remove the first two GCS layers, because DiffPool has an internal message-passing layer for the input features.
DiffPool and MinCut were trained in batch mode by zero-padding the adjacency and node attributes matrices.
All networks were trained using Adam with the default parameters of Keras, except for the learning rate. 

\noindent Hyperparameters:
\begin{itemize}
    \item Batch size: 8;
    \item Learning rate: 0.001;
    \item Weight decay: 0.00001;
    \item Epochs: models trained to convergence;
    \item Patience: 50 epochs;
    \item Repetitions per method and per dataset: 10;
    \item Data: we used the Benchmark Datasets for Graph Kernels as described in \cite{ivanov2019understanding}, that were modified to contain no isomorphic graphs. For each run, we randomly split the dataset and use 80\% of the data for training, 10\% for early stopping, and 10\% for testing.
\end{itemize}

\subsection{Graph regression}

\noindent We configure a GNN with the following structure: \textsc{ECC - ECC - GlobalPooling - Dense}, where \textsc{ECC} indicates an Edge-Conditioned Convolutional layer \cite{simonovsky2017dynamic} and \textsc{GlobalPooling} indicates the global pooling layer being tested. 
ECC layers have 32 units each, and ReLU activation. No regularization is applied to the GNN.
GAP is configured to use 32 units.
All networks were trained using Adam with the default parameters of Keras, except for the learning rate. We use the mean squared error as loss. 

Node features are one-hot encodings of the atomic number of each atom. Edge features are one-hot encodings of the bond type. The units of measurement for the target variables are: debye units (D) for Mu, a$_0^3$ ($a_0$ is the Bohr radius) for Alpha, and Hartree (Ha) for Homo and U0 \cite{ramakrishnan2014quantum}.

\noindent Hyperparameters:
\begin{itemize}
    \item Batch size: 32;
    \item Learning rate: 0.0005;
    \item Epochs: models trained to convergence;
    \item Patience: 10 epochs;
    \item Repetitions per method and per dataset: 5;
    \item Data: for each run, we randomly split the dataset and use 80\% of the molecules for training, 10\% for early stopping, and 10\% for testing.
\end{itemize}

\end{document}